\documentclass[letterpaper, 10 pt,conference]{ieeeconf}  

\IEEEoverridecommandlockouts                              

\usepackage{soul}
\usepackage{graphicx}
\usepackage{footnote,url}

\title{\LARGE \bf
Strawberry Detection Using a Heterogeneous Multi-Processor Platform}

\author{Samuel Brandenburg$^{1}$ and Pedro Machado$^{1,2}$ and Nikesh Lama$^{1}$ and T.M. McGinnity $^{1,3}$
\thanks{$^{1}$Samuel Brandenburg, Pedro Machado, Nikesh Lama, T.M. McGinnity are with Computational Neuroscience and Cognitive Robotics Group, 
School of Science and Technology, Nottingham Trent University, Clifton Campus,United Kingdom. 
        {\tt\small samuel.brandenbourg2016@my.ntu.ac.uk and  \{pedro.batpistamachado, nikesh.lama02, martin.McGinnity\}@ntu.ac.uk}}
\thanks{$^{2}$Pedro Machado with the Sundance Multiprocessor Technology Ltd., Chesham, Buckinghamshire, UK.  {\tt\small  pedro.m@sundance.com}}
\thanks{$^{3}$T.M. McGinnity is also with the Intelligent Systems Research Centre, Ulster University, Magee Campus, Northern Ireland, UK.  {\tt\small  tm.mcginnity@ulster.ac.uk}}
}

\begin{document}

\maketitle
\thispagestyle{empty}
\pagestyle{empty}

\begin{abstract}
Over the last few years, the number of precision farming projects has increased specifically in harvesting robots and many of which have made continued progress from identifying crops to grasping the desired fruit or vegetable. One of the most common issues found in precision farming projects is that successful application is heavily dependent not just on identifying the fruit but also on ensuring that localisation allows for accurate navigation. These issues become significant factors when the robot is not operating in a prearranged environment, or when vegetation becomes too thick, thus covering crop. Moreover, running a state-of-the-art deep learning algorithm on an embedded platform is also very challenging, resulting most of the times in low frame rates.
This paper proposes using the You Only Look Once version 3 (YOLOv3) Convolutional Neural Network (CNN) in combination with utilising image processing techniques for the application of precision farming robots targeting strawberry detection, accelerated on a heterogeneous multiprocessor platform. The results show a performance acceleration by five times when implemented on a Field-Programmable Gate Array (FPGA)  when compared with the same algorithm running on the processor side with an accuracy of 78.3\% over the test set comprised of 146 images.
\end{abstract}


\section{INTRODUCTION} \label{sec:introduction}

 
The majority of the UK's land is used for agriculture production \cite{Angus2009}.  The production of agricultural commodities aids in the effort to be self-sufficient as production sustains food levels in the country \cite{Angus2009}. In order to support such mass production, each year, the agriculture industry requires the recruitment of hundreds of skilled labours to operate in the country's farmland \cite{Bogue2020}. A substantial amount of workers are required given the size of the operation, and that most of the work comprises of manual labour. As such, manual workers spend a large portion of their day performing hours of repetitive task, such as picking fruit and vegetables. However, due to legislation changes in England, it has become more difficult for farmers to find the workforce they need for their fields \cite{Bogue2020}. It is important to note that the shortage of field workers was present before Brexit \cite{Maye2018}. Labour shortages ultimately increase the amount of food waste when crops are not collected before they spoil. Thus, shortening the amount of supply to a growing demand due to continued population growth \cite{White2018}. 

Several universities and private companies have invested in smart farming to help reduce labour shortages and increase the production of agricultural commodities \cite{Xiong2019}. While the automation of agriculture presents opportunities to counterbalance the labour shortage, smart farming also has the potential to be more effective than traditional farming equipment \cite{White2018}\cite{Dagar2018}.  For example, heavy machinery such as tractors, which is used in traditional farming, has been shown to decrease the number of crop yields due to soil compaction \cite{White2018}. In the UK Robotics and Autonomous Systems (RAS) white paper, it is argued that small intelligent farming robots can aid in the reduction of waste, improve the economy, decrease environmental harm and increase food sustainability \cite{White2018}. Therefore, there is a great need for innovation, that could be potentially met by using smart farming robots capable of performing at a commercial level. The soft fruit sector could especially benefit from automating harvest collection on high yield fruit like strawberries.

However, a robot capable of picking strawberries must meet certain requirements to be successful. The first requirement is detecting and localising the fruit. Detection and localisation are essential because it takes visual data as input. The collected visual information is then analysed using a variety of different methods. A standard and suitable design for detecting fruit use colour sensors that take the colours in the image and analyses the pixels by colour and shape. Calculations are conducted within the colour space to determine if there are any items in the picture. Once one or more strawberries are identified, a localisation strategy is needed. Localisation is the process of identifying the location of each located fruit. Once the fruit has been properly localised, trajectory planning is required. 

 Trajectory planning is the second essential task and equally as challenging as detecting the desired crop. For this, a pre-planning approach may be more suitable for reaching fruit located under challenging positions verse doing the detection, localisation and end-effector manipulation all at once. The project measures successful trajectory planning based on three main criteria's; starting with how well it can manoeuvre around without being blocked or entangled by vegetation. In the case where such criteria can not be met, the robot should proceed with the next detected strawberry. The second objective is to have the gripper positioned within grabbing distance of the desired fruit. The last objective measures robot control and perception by ensuring that nearby crops are not damaged.

The next task combines grasping the fruit and removing it from its peduncle. The challenges are to remove the peduncle and grasp the fruit without inflicting any damage to it. Removing the peduncle wrong leads to the fruit being exposed, causing early spoilage. The other crucial part of the task is grasping the strawberry with a suitable amount of pressure. Applying too much pressure can lead to bruising on the strawberry. Two methods to determine how much pressure to apply are the use of visual and pressure sensors. When a strawberry ripens, its' colour becomes darker; therefore,  the robot can begin adjusting what grip is suitable by the degree of ripeness. In the case of uneven ripeness, a pressure sensor on the gripper would be the most suitable means of determining the amount of pressure to apply. 

The overall smart farming process can be accomplished by improving farming vision, tactile sensing and perception. In this paper a vision system is achieved through the use of YOLOv3 operating on the Xilinx ZCU104 development board \footnote{Available online, \protect\url{https://www.xilinx.com/products/boards-and- kits/zcu104.html}, last accessed 30/07/2020}, a heterogeneous platform fitted with the Xilinx Multi-Processor ZU7EV which is analysed for its suitability for identifying and localising crops such as strawberries.

A brief literature review is presented in section \ref{sec:literature_review}, the methodology in presented in section \ref{sec:methodology}, preliminary results are discussed in section \ref{sec:preliminary_results} and the conclusions and future work are given in section \ref{sec:conclusion}.

\section{RELATED WORK} \label{sec:literature_review}
The UK employs up to 29,000 seasonal pickers and generates over 160,000 tons of fruit each year \cite{Xiong2020}. One can see that agriculture requires a tremendous amount of people to produce adequate returns. However, it is essential to note that while such a high number of people are employed there is still a shortage of workers. The shortage of workers is, in fact, an invaluable opportunity to maximise production through automating soft fruit harvesting like strawberries.
One significant aspect of automating the collection of strawberry is that they are grown in most places in the world \cite{Xiong2020}; thus, an automated harvest would bolster the efficiency of strawberry agriculture world-wide. Currently, there is an extended amount of research conducted on fruit and vegetable picking robots. These research efforts have brought about significant advances in fruit-picking robots, such as the Argobot\footnote{Available online, \protect\url{https://www.agrobot.com/}, last accessed: 31/07/2020}, which uses 24 end-effectors to pick strawberries, and Sweeper, which is a robot designed to collect yellow peppers \cite{Ostovar2018}. While there are many robots capable of locating and grasping said fruit or vegetable, none are at a commercial standard \cite{Xiong2019} as of yet. These robots have not yet achieved a better harvesting performance than the methods currently commercially used for fruit or vegetable collection \cite{Xiong2020}.  

Any robot task with crop picking is faced with the questions of what methods will the robot use for information gathering on its environment. For instance, will the system implement computer vision \cite{Hayashi2010}, and if so, what methods would it use to process visual information; would it use a neural network or a machine learning algorithm?  One group focused on designing an algorithm that used multi-template matching for identifying cucumbers in natural surroundings\cite{Bao2016}. Bao \textit{et al.} \cite{Bao2016} proposed the use of an image that serves as a reference point for incoming images. The reference point slides across an image and is then compared with the template to identify labels in picture \cite{Bao2016}. Their system used 65 templates to detect mature Radit cucumbers. The method proposed by Bao \textit{et al.} \cite{Bao2016} was designed to overcome the issue of misidentifying the cucumber due to its complex environment\cite{Bao2016}. The author states other methods which focus on the feature extraction of the cucumber such as the shape, or the colour, struggle with accurate detection. The cucumber, unlike the strawberry, is green, which makes it harder to differentiate between the vegetable and the leaf \cite{Bao2016}.
Nonetheless, the author reported that their implementation was able to achieve 98\% accuracy \cite{Bao2016}. In spite of the results being certainly interesting, the work was done in 2016, and new emerging methodologies should be explored. In this case, template matching can still be improved to distinguish crop despite occlusion and lighting\cite{Bogue2020}. However, template-matching shows promising results when differentiating between subtle differences.

Other projects have been done on creating a smart farming robot for apple extraction \cite{Ni2018}. Red apples, unlike cucumbers, are easier to distinguish from leaves. However, automating apple farming presents challenges in visually recognising maturity because of uneven maturity \cite{Ni2018}. The author proposed a visual method using red, green, and blue (RGB) sensors to identify mature apples.  One clear difference in the previous project on cucumber recognition was their visual processing method. They used a multi-template technique for visualisation. However, the Ni \textit{et al.} \cite{Ni2018} project utilised an image processing algorithm based on an RGB colour scheme. The colour scheme is popular for colour models because a large array of colours can be created from varying the degrees of red, green and blue values. The system was broken into separate tasks. The first task was acquiring visual data to process, which was done by taking snapshots of the apple tree.  Once the image was acquired, the image was then analysed based on its RGB traits \cite{Ni2018}. Image processing based on colours play a crucial role in identifying objects of interest from noise \cite{Ni2018}. Ni's method is beneficial because it is intuitive and able to adapt to changes in colour. However, it was asserted that Ni's method of intensifying the colour spectrum to identify apples might not be as effective in image processing if there are subtle differences with its environment\cite{Bao2016}. 
Furthermore, it may be beneficial to add supplemental vision techniques to increase the accuracy rate and reduce false positives. False positives mislead the system to believe it has the correct prediction \cite{Ni2018}. The most critical similarity is that apples and strawberries share the same colour. Therefore, image processing based on RGB sensors may prove essential for a foundational understanding of the farming environment for strawberries.

The red and green colour scheme for image processing was utilised in a smart farming project focused on automating cherry tomato harvesting \cite{Feng2018}. Similar to the previous projects, the main challenge was identifying mature cherry tomatoes \cite{Feng2018}. Feng \textit{et al.} \cite{Feng2018} also used a colour scheme method for image processing, but instead of using RGB, they used R-G. The author stated that a red and green model would help differentiate the targeted tomatoes from its background as a consequence of the colour intensity formed by having only red and green to represent their colour spectrum. The article reported having an 83\% harvest success, yet the robotic arm would sometimes collide with the branches, therefore, in its attempts to grab a tomato \cite{Feng2018}. Ultimately, the colour model can be used in different ways to reduce the amount of white noise in image processing \cite{Feng2018}.  The proposed method was reported useful for tomato picking, and it could be useful for collecting apples. It is unlikely that it would be useful in cucumber harvesting due to cucumbers sharing the same colour with their background. Intensifying the colour spectrum may only create more false positives identifying leaves as cucumbers. The proposed method could be beneficial for strawberry picking as they are sometimes grown in clusters. Strawberries also ripe unevenly, so identifying the mature strawberries could be more challenging when closely confined with other berries \cite{Xiong2019}.

Another group provided two research papers on strawberry picking. The first report used an RGB Depth camera and two infrared sensors to detect the depth of the object in its environment. The visual component of their robot system has two abilities: detection of individual strawberry, the localisation of the fruit for its separation. The process of filtering out pixel noise is used to assist in identifying strawberries by using a colour threshold\cite{Xiong2019}. The author stated that image processing based on RGB provides fast performance for live detection. It also discussed cases where the robot failed due to the vision system. They reported that their vision system was impaired by occlusions, duplicate detections, inaccurate localisation, and segmentation failures \cite{Xiong2019}.
Obstruction of the camera view appears to be the main issue in each smart farming project. It appears that it is more challenging when harvesting strawberries, probably, due to strawberries being small and easily covered by leaves and easily hidden from the robot's end-effector. Another unsuspected issue is duplicate detections of the same fruit. The research that covered cucumber detection used template-matching however, it appeared that objects did not obscure the camera view from taking sufficient images during testing. Nonetheless, a multi-template system may help mitigate duplicate detection. Xiong \textit{et al.} \cite{Xiong2020} attempted to improve their computer vision algorithm by creating a subsystem that created a threshold for adapting to colour changes specifically to sunlight. The subsystem did its calculations on 2-dimensional images and allowed for simplicity in their algorithm and fast performance. However, segmentation issues caused localisation error while processing the image. Xiong \textit{et al.} suggested that it could potentially be resolved by doing 3-dimensional image processing\cite{Xiong2020}. However, the approach would require more complex solutions, for instance, a deep learning algorithm \cite{Xiong2020}.

In addition to computer vision, it is important to consider what sensors can be used to gauge distance from the targeted crop. These are interlinked with the overall performance of the smart farming robot, and research shows that different sensors can be invaluable in simplifying distance calculations \cite{Bogue2020}. The use of different sensors is important because it can assist the trajectory planning for the chosen end-effector. Once the end-effector is designed, and its range of motion has been determined, a  trajectory planning algorithm can be tested \cite{Xiong2018}. All of these separate components working together are needed to plan a successful grasping attempt. An important note is how trajectory planning and design of the end-effector is based on the nature of the crop and raises important questions to answer when attempting a grasping application. Each robot is faced with visualising the targeted crop, using a well-designed end effector suited for grasping the harvest, and trajectory planning to the target crop. Previous work \cite{Xiong2018} shows that soft fruit requires more care when handling them, unlike apples, to prevent bruising. 

The size of the fruit and the way it grows may present additional difficulties when attempting to grasp the item. A good example is comparing how cucumbers grow compared to strawberries. Cucumbers grow in a way that makes them easy to reach because they usually are not obscured by vegetation. ; however, strawberries can grow in quite an unpredictable and hidden fashion\cite{Xiong2020}.  Therefore, it is essential first to understand the nature of how strawberries will grow and what planning can be done in advance to increase performance.  An essential aspect to consider when analysing the challenges to strawberry picking is their environment. Strawberries planted in greenhouses may grow in a more organised manner, so the fruits are less likely to be hidden from the view of the robot\cite{Ge2019}. As a result, it makes it easier for a robot to find and collect each strawberry. However, the same cannot be said about strawberries grown in a field, as they would require more probing around obstacles to find and grasp them\cite{Xiong2020}.  A low-efficiency rating will cost the agriculture industry unnecessary loss, thus defeating the purpose of smart farming and addressing the needs of the robot efficiently complete the tasks.

Another critical challenge is that strawberries ripe unevenly thus further complicating grasping the soft fruit. If the end-effector grasps the strawberry with too much pressure, the fruit can be damaged\cite{Hayashi2010}. Gauging the right amount of force must be adaptable to the ripeness of each fruit. One way to determine if a fruit is ready to collect is based on its ripeness which is reflected in its colour. The robot must adapt to the different ranges of red in strawberries to differentiate the pre-mature fruit from the mature ones. Finally, strawberries are small in size and sometimes grow in clusters. Strawberry picking is considerably challenging for a robot\cite{Xiong2020}, because the robot has to distinguish each fruit from the other while locating the mature strawberry in a cluster of different colours. Failure to accurately identify the right berry may result in picking the wrong one or damaging nearby strawberries. Therefore, such a task would not be feasible without computer vision.

\section{METHODOLOGY} \label{sec:methodology}
 Our review on automated robots clearly shows that while each picking task is different, the robot's visual sensory largely impacts its success. Although one could argue that the end-effector and gripping strategy plays an equal part in the success of precision farming, it is evident that without a reliable and robust way to visually survey the crops environment, the following task may fail. Thus the contribution to our paper analyses how YOLOv3 operating on the Xilinx ZCU104 development board could improve strawberry classification—thus providing an essential function for fruit picking robots.

A fruit grasping algorithm includes five distinct phases, namely, 1) fruit classification: use the CNN algorithm to classify fruits in the image; 2) pose estimation: estimate the best pose to pick the object; 3) path searching: search the best path assuming the current position of the objects in the scene; 4) path planning: predict moving objects trajectory and estimate the optimal path while avoiding collisions and 5) grasping: process grasping the target fruit. 

\begin{figure}[htb!]
\includegraphics[width=0.25\textwidth]{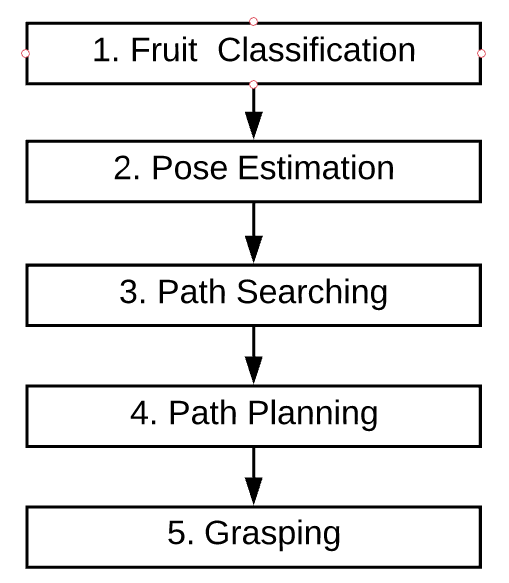}
\centering
  \caption{Fruit grasping steps} \label{fig:fruit_grasping}
\end{figure}

The fruit grasping steps are represented in Figure~\ref{fig:fruit_grasping}.

The focus of this paper is on phase 1: the fruit classification. The objective during the classification stage is to improve strawberry detection in 6 key areas. These areas comprise of improving localisation accuracy, building model durability to light variances, enabling background adaptability, identifying partially hidden strawberries, increasing accuracy confidence, and achieve real-time processing speed. To accomplish such feats, the project utilises the deep learning network YOLOV3, which will operate on a Xilinx ZCU104 development board.

Our custom YOLOV3 model was trained on 1454 images of strawberries obtain from various sources. The selection of images played a key part in ensuring our model could adapt to light and background variances. This is also important when training the network to identify strawberries partially engulfed by vegetation. To reduce the resource demands for training, the images were resized to 416x416. Reducing the size of the images allowed our machines to process training better, but it also meant that there was a loss of detail in the images.  The model was trained up to 30k epoch's \footnote{One epoch is when an entire dataset is passed forward and backwards through the neural network only once.} which iterated by 10k. After testing each model trained at varying epochs, it was evaluated that the model trained at 20k epochs produced the best accuracy; Therefore, this model was chosen to operate on the Xilinx ZCU104. 

The Xilinx ZCU104 development board, powered by a powerful Xilinx Zynq UltraScale+ ZU7EV MPSoC, was used to implement and accelerate the YOLOv3 at the edge. The Xilinx Zynq UltraScale+ ZU7EV MPSoC is characterised by including a 64-bit quad-core ARM Cortex A53 as the Processor System (PS) for running a standard embedded Linux Operating System, a Graphical Processing Unit (GPU) for accelerating the graphics processing, Field-Programmable Gate Arrays (FPGA) for accelerating the state-of-the-art AI algorithms and Real-Time Processing Unit (RTU) for processing events in real-time\footnote{Available online, \protect\url{https://www.xilinx.com/products/silicon-devices/soc/zynq-ultrascale-mpsoc.html}, last accessed 31/07/2020}. Only the PS and FPGA units were used.

\begin{figure*}[htb!]
\includegraphics[width=0.8\textwidth]{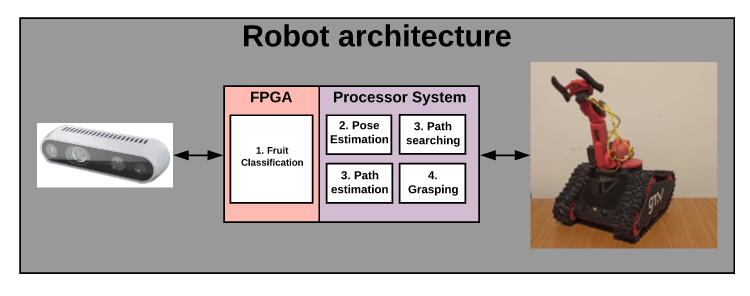}
\centering
  \caption{High Level Block Diagram} \label{fig:high-level_architecture}
\end{figure*}

An Intel RealSense D435i camera\footnote{Available online, \protect\url{https://www.intelrealsense.com/depth-camera-d435i/},last accessed 31/07/2020} is used to capture RGB and depth image frames that are forwarded to the MPSoC. The classification is accelerated on the FPGA side, and the remaining steps are completed on the processor system, which then controls the mechanical arm to pick the fruit.
The overall approach is shown in Figure~\ref{fig:high-level_architecture}.

Resource consumption in embedded systems and the lack of fruit and vegetable datasets are two drawbacks with using deep learning for crop classification. Thus the reasons why traditional image processing techniques have been widely used\cite{OMahony2020}. However, deep learning algorithms used for object detection are more adaptable to variance in light and are able to detect sophisticated patterns in data. While using deep learning for precision farming trades off speed for recognising complex images, utilising an optimised AI model on a Multi-Processor System-on-Chip (MPSoC) could help balance the scale between speed and accuracy. To do this, the Vitis-AI tool was used to quantise the 32-bit floating-point model into an 8-bit processing model. Quantising the model reduces its overall complexity while ensuring very little accuracy is lost
. The quantising process is then followed up by compiling the model which enables the model to utilise the board's deep processing unit (DPU). 

The Xilinx Deep Learning Processor Unit (DPU) is a programmable engine that was designed for accelerating convolutional neural networks. The DPU contains a register module, a data controller module, and a convolution computing module. There is a specialised instruction set for the DPU, which enables it to accelerate several state-of-the-art convolutional neural networks (CNNs)\footnote{Available online, \protect\url{https://www.xilinx.com/products/intellectual-property/dpu.html}, last accessed 31/07/2020}.


The DPU was configured with the Xilinx recommended parameters\footnote{Available online, \protect\url{https://www.xilinx.com/support/documentation/ip_documentation/dpu/v3_2/pg338-dpu.pdf}, last accessed: 30/07/2020}.

\begin{figure}[htb!]
\includegraphics[width=0.48\textwidth]{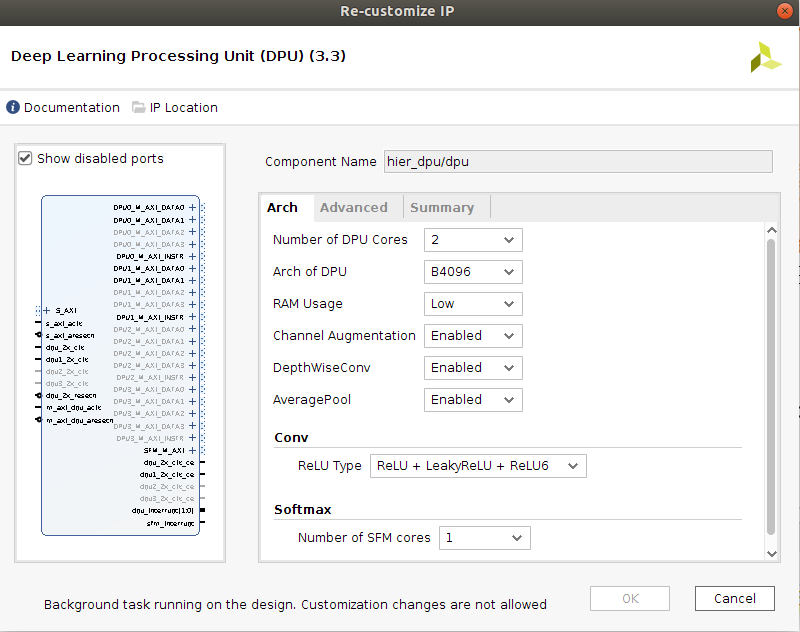}
\centering
  \caption{DPU configuration with 2 cores of the B4096 architecture, low RAM usage, channel augmentation, Depth Wise convolution, ReLU type: ReLU + LeakyReLU + ReLU6 and 1 SFM core}\label{fig:dpu_conf}
\end{figure}

The Vitis-AI development environment was designed for AI inference targeting the Xilinx hardware platforms, the including ZCU104 used in this project. It consists of optimised IP (e.g. the DPU), tools, libraries, models, and example designs\footnote{
Available online, \protect\url{https://www.xilinx.com/products/design-tools/vitis/vitis-ai.html}, last accessed 31/07/2020}. 

Table~\ref{tab:source of information} shows a list of resources and tools used in the project to perform the classification using YOLOv3 accelerated on the FPGA side.
\begin{table}[htb!] \label{tab:source of information} \caption{List of equipment and tools}
\begin{tabular}{|p{2cm}|p{2cm}|p{3.5cm}|} 
\hline
\textbf{Resources} & \textbf{Type} & \textbf{Description}  \\ \hline
\hline
GMV/Sundance Dennis robot prototype & Dennis robot prototype fitted with the Sundance VCS-junior system prototype & The Sundance VCS-junior system prototype is light version of the Xilinx ZCU104 designed to control small robots.  \\ \hline
Xilinx ZCU104 development board & Xilinx Zynq UltraScale+ ZU7EV MPSoC & FPGA used for accelerating the classification of strawberries.  \\ \hline
Intel RealSense D435i sensor & Camera & Camera used to capture colour and depth image frames.  \\ \hline
YOLOv3 algorithm & AI-Model & A single shot algorithm that localises and classifies objects from one input.  \\ \hline
Xilinx Vivado & Xilinx EDA tool & used for implementing the DPU on the FPGA side and connect it to the PS side using the AXI bus.  \\ \hline
Xilinx DPU & Xilinx Deep Processing Unit & used for accelerating YOLOv3 on the FPGA side.  \\ \hline
Vitis-AI quantizer & Xilinx tool & Converts 32-bit floating-point weight into a fix-point 8 bit integer    \\ \cline{1-3}
Vitis-AI compiler & Xilinx tool & Plots the AI model to an instruction set to be used by the DPU  \\ \cline{1-3}
Deep Learning processor unit & Xilinx tool & An optimised engine which can be programmed for deep neural networks use.  \\ \hline
\end{tabular}%
\end{table}

The report generated by Xilinx Vivado shows the utilisation of:
\begin{itemize}
    \item 40\% of Look-up-Tables (LUT): used to buffer data
    \item 8\% of LUT ditributed RAM (LUTRAM): used as small data buffers;
    \item 37\% of flip-flops (FF): used to describe logic circuits;
    \item 57\% of built-in RAM (BRAM): store data;
    \item 55\% of Digital Signal Processing (DSP): used to process signals inside the FPGA;
    \item 1\% of Global Clock Buffer (BUFG): used to buffer signal and assure that all the logic circuits receive the same clock signal within the acceptable tolerances;
    \item 13\% of Phase Locked Loop (PLL): clock controller is used compensate the clock signal; 
\end{itemize}

The resources utilisation shows that about half of the FPGA side is available to perform acceleration of algorithms from the next phases of fruit picking (i.e. pose estimation until grasping).

\section{PRELIMINARY RESULTS} \label{sec:preliminary_results}


Two test were conducted to determine how well the model could identify strawberries in a farm setting. The first test assessed the detection percentage over the validation set of a 146 images. The validation set contained images of strawberries not limited to a farming environment. When testing the model threshold was set to 70\% this parameter reduces the risk of false positives. As a result, the model was able to detect at an impressive 78.3\% over the validation set.  The second test used a youtube video collected on a strawberry farm\footnote{Available online,\protect\url{https://www.youtube.com/watch?v=-br6dVv9yDs}, last accessed 15/08/2020}. It is important to note that the full video does not contain strawberries on frame; however, this allowed us to gather insight on how well the model would perform in adverse environments. The video contained a total of 7439 frames. Our experiment analysed the video and created two filters. The first was to check how many frames did our model identify a potential strawberry. In this case, the model detected strawberries in 3371 frames.  Of this, 2346 frames were identified as having a low confidence score ranging from .30 percent to .60 percent. When identifying these frames, it was evident that the model, even at a low confidence rate had few false positives; however,  the model was not able to capture a high degree of all the strawberries in any frame. Furthermore, the network struggled to identify closely coupled strawberries. Another key area was localisation accuracy; this area focuses on how well was the strawberry encapsulated in the bounding box. Unfortunately, results were volatile; therefore, measuring the localisation accuracy was unreliable. Certain frames showed that the strawberry was located within the centre of the bound box, while other images of strawberries were partially encapsulated. In some cases, the network would falsely classify the ground or a leaf with red coating. Due to the current localisation results, other methods such as segmentation by colour should be used to calibrate the bounding box and help reduce false positives. The second metrics to evaluate was how fast the network detected on the ZCU104 Development board after being acclerated by the DPU.

\begin{table}[htb!] \label{table:benchmark results} \caption{Frame rates obtained running YOLOv3 in different architectures}
\begin{tabular}{|p{2cm}|p{1.5cm}|p{4cm}|} 
\hline
\textbf{Model} & \textbf{Frames per second (FPS) } & \textbf{Device}  \\ \hline
\hline
YOLOv3 on the processor system & 2.66 & Raspberry PI 3 model B \\\hline
YOLOv3 accelerated on the FPGA & 13.05 & ZCU104  with no modifications \\\hline
YOLOv3 \cite{Redmon2018}  & 30 & Workstation fitted with a NVIDIA Quadro P2200 and Intel Xeon Gold with 20 hyper-threaded cores and 64 GB of DDR3\\\hline 
\end{tabular}
\end{table}

\begin{figure}[htb!]
\includegraphics[width=0.2\textwidth]{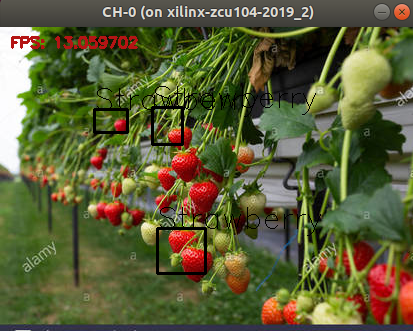}
\centering
  \caption{Classification of a strawberry using the YOLOv3 running on the MPSoC} \label{fig:classification}
\end{figure}

Table~\ref{table:benchmark results} displays the frames per second achieved running YOLOV3 on different processors. The YOLOv3 achieved 30 FPS when running on the powerful workstation and accelerated by a powerful Graphics Processing Unit (GPU) which is not suitable to be installed on robots and has more processing resources than embedded processors such as the 64-bit quad-core ARM cortex a53 which is found in the ZCU104. When running the YOLOv3 on the Xilinx ZU7EV Processor System, 2.66 FPS was achieved whereas, the ZCU104 Processor System accelerated by the DPU described on the FPGA side achieved the impressive 13.05 FPS\ref{fig:classification} demonstrated when detecting a strawberry. It is important to highlight that the maximum power consumption was 20W as opposed to the typical 350W consumed by the workstation described in Table~\ref{table:benchmark results}.

The network performed surprisingly well when identifying strawberries in different lighting. While true, the networks detection on the bases of lighting was not consistent. The network tended to identify strawberries that were well lighted better than the ones cascaded by a shadow. 

 In this evaluation, the model showed that different backgrounds did not affect the network. This could be seen in the video as the background was everchanging. Another key aspect is that the video was not used in the training data.

Another property that displayed that the model began to generalise is that it showed signs that it was capable of detecting partially hidden strawberries.  Generalisation was displayed at different ranges of distances proving that the model was beginning to become quite robust. 
Furthermore, our model was able to detect partially hidden strawberries. While the model was able to identify strawberries hidden in vegetation, there is also the issue of false-positive.

\begin{figure}[htb!]
\includegraphics[width=0.2\textwidth]{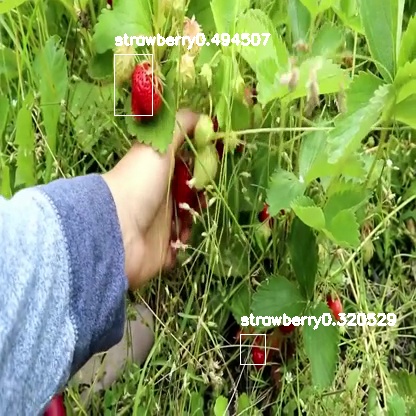}
\includegraphics[width=0.2\textwidth]{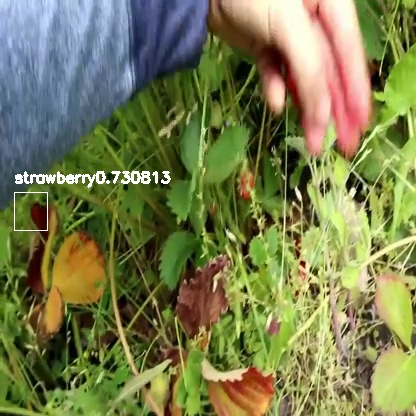}
\includegraphics[width=0.2\textwidth]{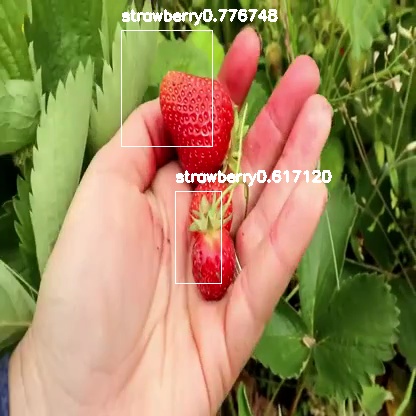}
\includegraphics[width=0.2\textwidth]{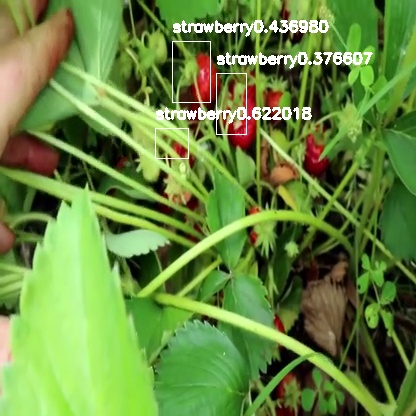}
\centering
  \caption{Classification of strawberries over four image frames extracted from a video footage} \label{fig:Farmclassification}
\end{figure}

\section{CONCLUSION AND FUTURE WORK} \label{sec:conclusion}

Currently, the project is proceeding to the second development stage, which poses estimation, while improving classification by increasing the training dataset. Therefore, the current achievements of this work is the classification of strawberries using YOLOV3 on a ZCU104 development board which is accelerated using a DPU. Hence, achieving fast and accurate detection at 13.05 FPS  within its' testing stage. However, as with many deep learning algorithms, their accuracy is dependent on the quality and volume of data. Thus,  there is still a need to improve the accuracy of the classifications with our model. This especially true when detecting strawberries in complex environments like greenhouses or farmland. Nonetheless, the current results show that our model can adapt to varying degrees of light and can accurately detect strawberries in ideal positions. The key takes away is to include more data, especially images that contain strawberries in clusters. As such, it is expected that problems will arise when testing the algorithm in different settings—another foreseeable issue adapting the algorithm to detect strawberries in clusters. More research and testing would need to be conducted to find a potential solution.

 Future work will include increasing the model's FPS. This will be done by utilising the Xilinx optimiser tool to prune and finetune the model. Most neural networks have redundant layers that are not essentially needed. The Xilinx optimiser tool allows an engineer to remove specified layers to improve detection time. The removal of layers, however, does reduce the models' accuracy slightly, but finetuning the model can help restore accuracy. Thus, achieving faster detection with little accuracy lost. Future work will also include using a colour segmentation algorithm to help calibrate the bounding boxes over the strawberry and assist with estimating the position of the strawberry. Once these key conditions have been met, further testing will include the use of a robot. However, our robot will accomplish faster image processing by accelerating the AI model using the Sundance VCS-1\footnote{Available online, \protect\url{https://www.sundance.com/vcs-1/}, last accessed 30/07/2020} and VCS-junior systems fitted with a powerful Xilinx ZU4EV MPSoC \footnote{Available online,\protect\url{https://www.xilinx.com/support/documentation/selection-guides/zynq-ultrascale-plus-product-selection-guide.pdf}, last accessed 30/07/2020}.

\addtolength{\textheight}{-12cm}   




\section*{ACKNOWLEDGMENT}
The authors would like to thank to Sundance Multiprocessor Technology Ltd. and the Sundance University Program for the financial support, access to equipment, and technical support provided.


\bibliographystyle{unsrt}
\bibliography{ref.bib}

\end{document}